\documentclass{article}
\usepackage{style}

\usepackage{times}
\usepackage{soul}
\usepackage{url}
\usepackage[hidelinks]{hyperref}
\usepackage[utf8]{inputenc}
\usepackage[small]{caption}
\usepackage{graphicx}
\usepackage{amsmath}
\usepackage{amsthm}
\usepackage{booktabs}
\usepackage{algorithm}
\usepackage{algorithmic}
\usepackage[switch]{lineno}
\usepackage{subcaption}
\usepackage{xcolor}
\usepackage{amsfonts}
\usepackage{multirow}
\usepackage{xcolor}
\usepackage{colortbl}
\usepackage{hyperref}
\usepackage{natbib}

\title{Large Language Models for Time Series: A Survey}

\author{
Xiyuan Zhang
\and
Ranak Roy Chowdhury\and
Rajesh K. Gupta\And
Jingbo Shang\\
\affiliations
University of California, San Diego\\
\emails
\{xiyuanzh, rrchowdh, rgupta, jshang\}@ucsd.edu
}

\begin{document}

\maketitle

\begin{abstract}
    Large Language Models (LLMs) have seen significant use in domains such as natural language processing and computer vision. Going beyond text, image and graphics, LLMs present a significant potential for analysis of time series data, benefiting domains such as climate, IoT, healthcare, traffic, audio and finance. This survey paper provides an in-depth exploration and a detailed taxonomy of the various methodologies employed to harness the power of LLMs for time series analysis. We address the inherent challenge of bridging the gap between LLMs' original text data training and the numerical nature of time series data, and explore strategies for transferring and distilling knowledge from LLMs to numerical time series analysis. 
    We detail various methodologies, including (1) direct prompting of LLMs, (2) time series quantization, (3) aligning techniques, (4) utilization of the vision modality as a bridging mechanism, and (5) the combination of LLMs with tools. Additionally, this survey offers a comprehensive overview of the existing multimodal time series and text datasets and delves into the challenges and future opportunities of this emerging field. We maintain an up-to-date Github repository\footnote{\url{https://github.com/xiyuanzh/awesome-llm-time-series}} which includes all the papers and datasets discussed in the survey.
\end{abstract}
\section{Introduction}

Time series analysis plays a critical role in a variety of fields, including climate modeling, traffic management, healthcare monitoring and finance analytics. Time series analysis comprises a wide range of tasks such as classification~\citep{liu2023large}, forecasting~\citep{gruver2023large}, anomaly detection~\citep{zhang2023large}, and imputation~\citep{chen2023gatgpt}. Traditionally, these tasks have been tackled using classical signal processing techniques such as frequency analysis and decomposition-based approaches. More recently, deep learning approaches like Convolutional Neural Networks (CNNs), Long Short-Term Memory networks (LSTMs), and Transformers~\citep{wen2022transformers} have revolutionized this field and proved effective in extracting meaningful patterns from time series data, making them the primary approaches of time series analysis in various domains.

\begin{figure}[t]
    \includegraphics[width=0.48\textwidth]{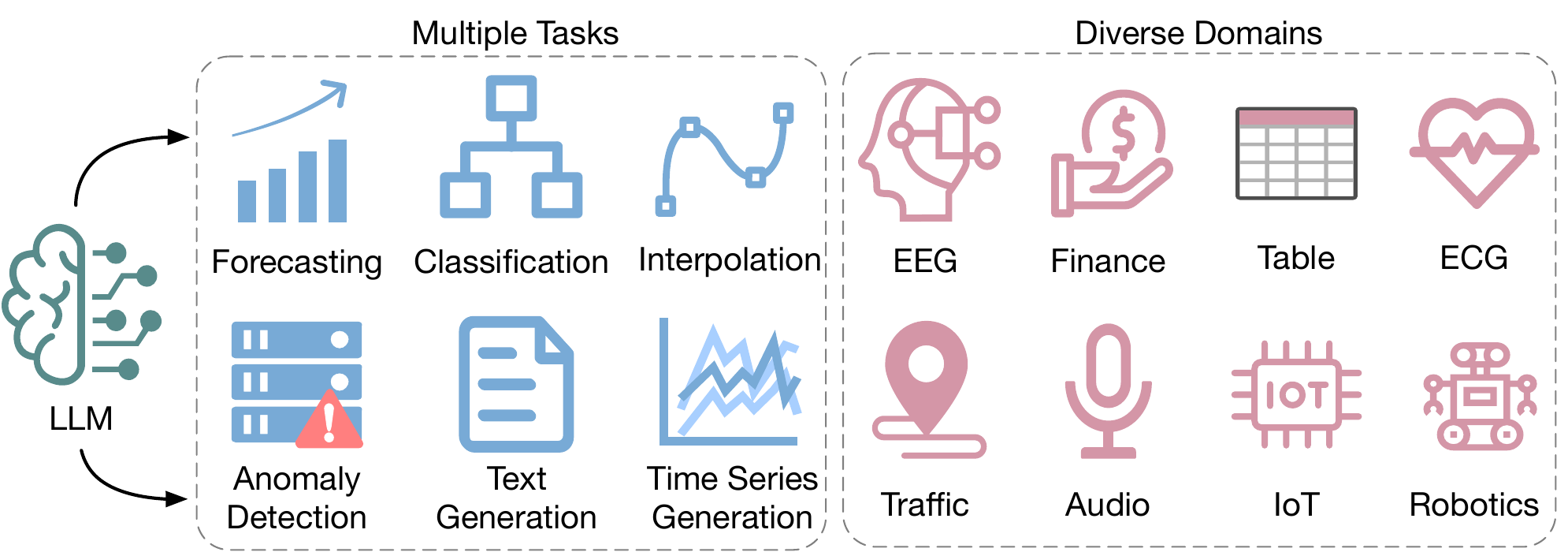}
    \caption{Large language models have recently been applied for various time series tasks in diverse application domains.}
    \label{fig:app}
\end{figure}

\begin{figure*}[t]
    \includegraphics[width=\textwidth]{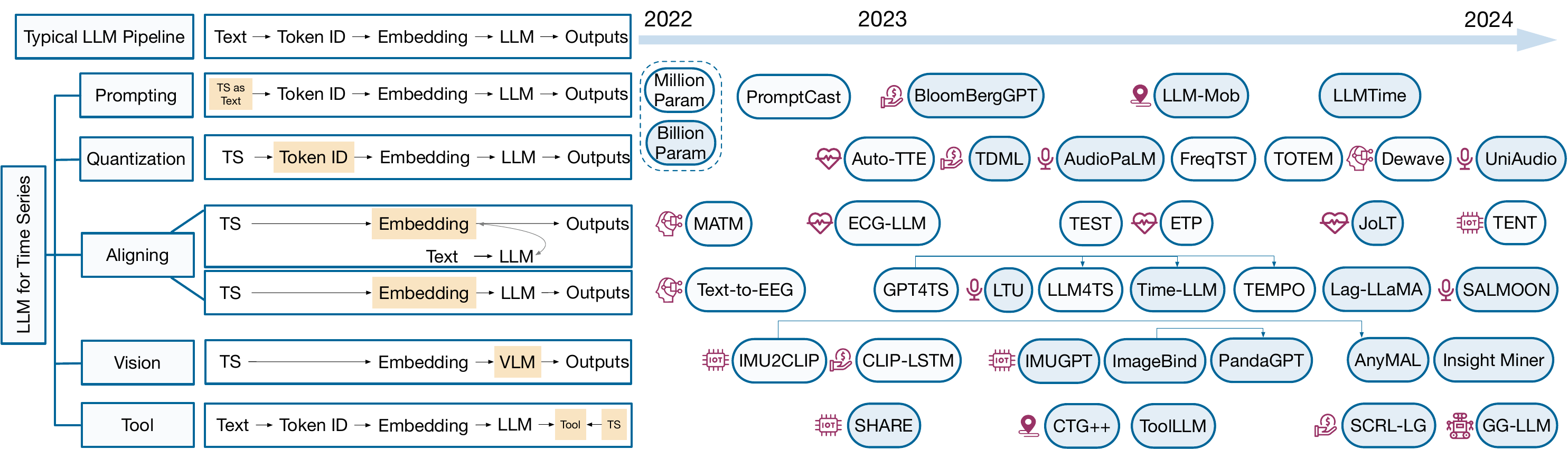}
    \caption{Left: Taxonomy of LLMs for time series analysis (prompting, quantization, aligning which is further categorized into two groups as detailed in Figure~\ref{fig:alignment}, vision as bridge, tool integration). For each category, key distinctions are drawn in comparison to the standard LLM pipeline shown at the top of the figure. Right: We present representative works for each category, sorted by their publication dates. The use of arrows indicates that later works build upon earlier studies. Dark(light)-colored boxes represent billion(million)-parameter models. Icons to the left of the text boxes represent the application domains of domain-specific models, with icons' meanings illustrated in Figure~\ref{fig:app}.
    }
    \label{fig:taxo}
\end{figure*}

In recent years, Large Language Models (LLMs) have gained substantial attention particularly in the fields of Natural Language Processing (NLP) and Computer Vision (CV). Prominent models such as GPT-4~\citep{OpenAI2023GPT4TR} have transformed the landscape of text processing by offering unprecedented accuracy in tasks such as text generation, translation, sentiment analysis, question answering and summarization. In the CV domain, Large Multimodal Models (LMMs) have also facilitated advancements in image recognition, object detection, and generative tasks, leading to more intelligent and capable visual systems~\citep{song2023bridge}. Inspired by these successes, researchers are now exploring the potential of LLMs in the realm of time series analysis, expecting further breakthroughs, as shown in Figure~\ref{fig:app}. While several surveys offer a broad perspective on large models for time series in general~\citep{jin2023large,ma2023survey}, these do not specifically focus on LLMs or the key challenge of bridging modality gap, which stems from LLMs being originally trained on discrete textual data, in contrast to the continuous numerical nature of time series.

Our survey uniquely contributes to the existing literature by emphasizing how to bridge such modality gap and transfer knowledge from LLMs for time series analysis. Our survey also covers more diverse application domains, ranging from climate, Internet of Things (IoT), to healthcare, traffic management, and finance. Moreover, certain intrinsic properties of time series, like continuity, auto-regressiveness, and dependency on the sampling rate, are also shared by audio, speech, and music data. Therefore, we also present representative LLM-based works from these domains to explore how we can use LLMs for other types of time series. 
We present a comprehensive taxonomy by categorizing these methodologies into five distinct groups, as shown in Figure~\ref{fig:taxo}. If we outline typical LLM-driven NLP pipelines in five stages - input text, tokenization, embedding, LLM, output - then each category of our taxonomy targets one specific stage in this pipeline. Specifically, (i) \emph{Prompting} (input stage) treats time series data as raw text and directly prompts LLMs with time series; (ii) \emph{Time Series Quantization} (tokenization stage) discretizes time series as special tokens for LLMs to process; (iii) \emph{Aligning} (embedding stage) designs time series encoder to align time series embeddings with language space; (iv) \emph{Vision as Bridge} (LLM stage) connects time series with Vision-Lanuage Models (VLM) by employing visual representations as a bridge; (v) \emph{Tool Integration} (output stage) adopts language models to output tools to benefit time series analysis. Beyond this taxonomy, our survey also compiles an extensive list of existing multimodal datasets that incorporate both time series and text. We conclude our paper by discussing future research directions in this emerging and promising field.
\section{Background and Problem Formulation}

Large language models are characterized by their vast number of parameters and extensive training data. They excel in understanding, generating, and interpreting human language and recently represent a significant advancement in artificial intelligence. The inception of LLMs can be traced back to models like GPT-2~\citep{radford2019language}, BERT~\citep{devlin2018bert}, BART~\citep{lewis2019bart}, and T5~\citep{raffel2020exploring}, which laid the foundational architecture. Over time, the evolution of these models has been marked by increasing complexity and capabilities, such as LLAMA-2~\citep{touvron2023llama}, PaLM~\citep{chowdhery2023palm}, and GPT-4. More recently, researchers have developed multimodal large language models to integrate and interpret multiple forms of data, such as text, images, and time series, to achieve a more comprehensive understanding of information.  

This survey focuses on how LLMs could benefit time series analysis. We first define the mathematical formulation for the input and output, which may contain time series or (and) text depending on the downstream tasks, as well as the models.

\noindent \textbf{Input}: denoted as $\mathbf{x}$, composed of time series $\mathbf{x}_s \in \mathbb{R}^{T \times c}$ and optional text data $\mathbf{x}_t$ represented as strings, where $T,c$ represent the sequence length and the number of features.

\noindent \textbf{Output}: denoted as $\mathbf{y}$ and may represent time series, text or numbers depending on the specific downstream task. For time series generation or forecasting task, $\mathbf{y}$ represents generated time series $\mathbf{y}_s$ or predicted $k$-step future time series $\mathbf{y}_s^{T+1:T+k}$. For text generation task, such as report generation, $\mathbf{y}$ represents text data $\mathbf{y}_t$. For time series classification or regression task, $\mathbf{y}$ represents numbers indicating the predicted classes or numerical values. 

\noindent \textbf{Model}: We use $f_{\theta}$ parameterized by $\theta$, $g_{\phi}$ parameterized by $\phi$, and $h_{\psi}$ parameterized by $\psi$ to represent language, time series and vision models, where $f_{\theta}$ is typically initialized from pre-trained large language models. We optimize parameters $\theta$, $\phi$ and $\psi$ through loss function $\mathcal{L}$.

\section{Taxonomy}
In this section, we detail our taxonomy of applying LLMs for time series analysis, categorized by five groups. We summarize the representative works, mathematical formulation, advantages and limitations of each category in Table~\ref{tab:summary}.

\subsection{Prompting}

\renewcommand{\arraystretch}{1.5}
\begin{table*}[t]
\caption{Examples of representative direct prompting methods.}
\centering
{\small
\setlength{\tabcolsep}{1mm}{
\scalebox{0.95}{
\begin{tabular}{c|l}
\toprule
\rowcolor{gray!20} \textbf{Method} & \textbf{Example}  \\ \midrule
\multirow{2}{*}{PromptCast~\citep{xue2022promptcast}} & ``From $\{t_1\}$ to $\{t_\mathrm{obs}\}$, the average temperature of region $\{U_m\}$ was $\{x^m_t\}$ degree on each day. What is the  \\ 
& temperature going to be on $\{t_\mathrm{obs}\}$?'' \\ \hline
\multirow{2}{*}{\citet{liu2023large}} & ``Classify the following accelerometer data in meters per second squared as either walking or running: \\
& 0.052,0.052,0.052,0.051,0.052,0.055,0.051,0.056,0.06,0.064'' \\ \hline
\multirow{2}{*}{TabLLM~\citep{hegselmann2023tabllm}} & ``The person is 42 years old
and has a Master’s degree. She gained \$594. Does this person earn more than \\ 
& 50000 dollars? Yes or no? Answer:'' \\ \hline
LLMTime~\citep{gruver2023large} & ``0.123, 1.23, 12.3, 123.0'' $\rightarrow$ ``1 2 , 1 2 3 , 1 2 3 0 , 1 2 3 0 0'' \\
\bottomrule
\end{tabular}}}}
\label{tab:prompt}
\end{table*}

\textbf{Number-Agnostic Tokenization}: The method treats numerical time series as raw textual data and directly prompts existing LLMs. For example, PromptCast~\citep{xue2022promptcast} proposes prompt-based time series forecasting by converting numerical time series into text prompts and forecasting time series in a sentence-to-sentence manner. The input prompts are composed of context and questions following pre-defined templates. An illustrative prompt template for temperature forecasting, along with examples from other representative works, are showcased in Table~\ref{tab:prompt}. 
Similar prompting methods have been applied to forecast Place-of-Interest (POI) customer flows (AuxMobLCast~\citep{xue2022leveraging}), energy load~\citep{xue2023utilizing}, and user's next location (LLM-Mob~\citep{wang2023would}). 
\citet{liu2023large} prompt PaLM-24B for health-related tasks such as activity recognition and daily stress estimate.
TabLLM~\citep{hegselmann2023tabllm} prompts large language models with a serialization of the tabular data to a natural-language string for few-shot and zero-shot tabular data classification. \citet{zhang2023large} prompt large language models to detect anomalous behaviors from mobility data. \citet{xie2023wall} extract historical price features such as open, close, high, and low prices to prompt ChatGPT in a zero-shot fashion.  

\textbf{Number-Specific Tokenization}: More recently, LLMTime~\citep{gruver2023large} pointed out that Byte Pair Encoding (BPE) tokenization has the limitation of breaking a single number into tokens that don’t align with the digits, leading to inconsistent tokenization across different floating point numbers and complicating arithmetic operations~\citep{spathis2023first}. Therefore, following LLMs such as LLaMA and PaLM, they propose to insert spaces between digits to ensure distinct tokenization of each digit and use a comma (``,'') to separate each time step in a time series. They also scale time series to optimize token usage and keep fixed precision (e.g., two digits of precision) to efficiently manage context length. Meanwhile, BloomberGPT~\citep{wu2023bloomberggpt} trains on financial data with text and numerical data and places each digit in its own chunk to better handle numbers. Using similar space-prefixed tokenization, \citet{mirchandani2023large} show that LLMs are general pattern machines capable of sequence transformation, completion and improvement. 

\subsection{Quantization}

\begin{figure}[t]
      \centering
      \begin{subfigure}[b]{0.45\textwidth}
          \centering
          \includegraphics[width=\textwidth]{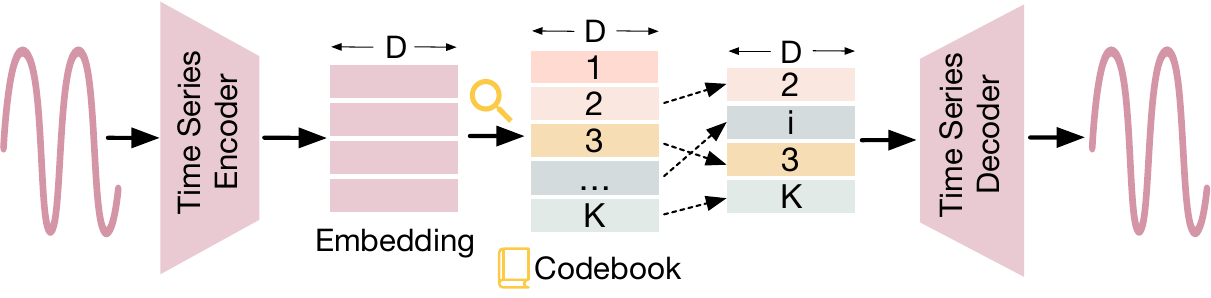}
          \caption{VQ-VAE based quantization method.}
          \label{fig:vq-vae}
      \end{subfigure}
      \begin{subfigure}[b]{0.45\textwidth}
          \centering
          \includegraphics[width=\textwidth]{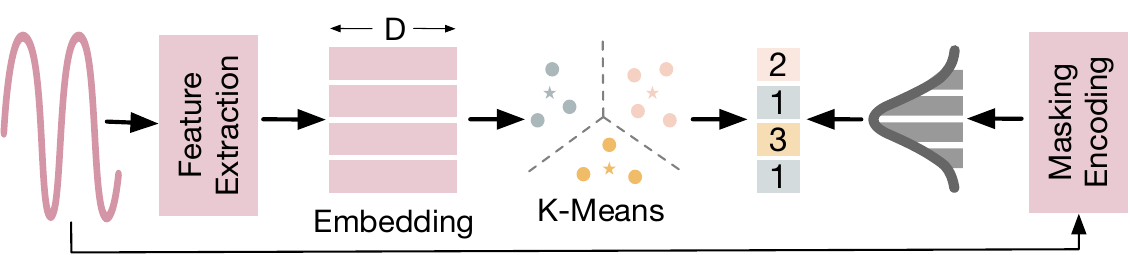}
          \caption{K-Means based quantization method.}
          \label{fig:k-means}
      \end{subfigure}
    \caption{Two types of index-based quantization methods.}
    \label{fig:quantize}
\end{figure}

Quantization based method~\citep{rabanser2020effectiveness} converts numerical data into discrete representations as input to LLMs. This approach can be further divided into two main categories based on the discretization technique employed.

\textbf{Discrete Indices from VQ-VAE}: The first type of quantization method transforms continuous time series into discrete indices as tokens. Among them one of the most popular methods is training a Vector Quantized-Variational AutoEncoder (VQ-VAE)~\citep{van2017neural}, which learns a codebook $\mathcal{C}=\{\mathbf{c}_i\}_{i=1}^K$ of $K$ $D$-dimensional codewords $\mathbf{c}_i \in \mathbb{R}^D$ to capture the latent representations, as illustrated in Figure~\ref{fig:vq-vae}. The method identifies the nearest neighbor $k_i$ of each step $i$ of the encoded time series representation $g_{\phi}(\mathbf{x}_s) \in \mathbb{R}^{\frac{T}{S} \times D}$ in the codebook ($S$ denotes the cumulative stride of VQ-VAE encoder), and uses the corresponding indices $\mathbf{k}$ as the quantized input to language models: 
\begin{equation}
    \mathbf{q}_i = \mathbf{c}_{k_i}, k_i = \arg\min_j \| g_{\phi}(\mathbf{x}_s)_i - \mathbf{c}_j \|_2, \mathbf{k} = [k_i]_{i=1}^{\frac{T}{S}}.
\end{equation}

Based on VQ-VAE, Auto-TTE~\citep{chung2023text} quantizes ECGs into discrete formats and generates 12-lead ECG signals conditioned on text reports. DeWave~\citep{duan2023dewave} adapts VQ-VAE to derive discrete codex encoding and aligns it with pre-trained BART for open-vocabulary EEG-to-text translation tasks. 
TOTEM~\citep{anonymous2023time} also quantizes time series through VQ-VAE as input to Transformers for multiple downstream applications such as forecasting, classification, and translation. 
In the audio domain, UniAudio~\citep{yang2023uniaudio} tokenizes different types of target audio using Residual Vector Quantization (RVQ)~\citep{zeghidour2021soundstream} (a hierarchy of multiple vector quantizers) and supports 11 audio generation tasks.
VioLA~\citep{wang2023viola} unifies various crossmodal tasks involving speech and text by converting speech utterances to discrete tokens through RVQ.
AudioGen~\citep{kreuk2022audiogen} learns discrete audio representations using vector quantization layers and generates audio samples conditioned on text inputs. 

\textbf{Discrete Indices from K-Means}: Apart from employing VQ-VAE, researchers have also explored K-Means clustering for index-based tokenization, which uses the centroid indices as discretized tokens~\citep{hsu2021hubert}, as shown in Figure~\ref{fig:k-means}. Such methods are mostly applied in the audio domain. For example,  SpeechGPT~\citep{zhang2023speechgpt} shows capability to perceive and generate multi-modal contents using K-Means based discrete unit extractor. AudioLM~\citep{borsos2023audiolm} discretizes codes produced by a neural audio codec using K-means clustering to achieve high-quality synthesis. It also combines discretized activations of language models pre-trained on audio using RVQ to capture long-term structure. Following the same quantization procedure, AudioPaLM~\citep{rubenstein2023audiopalm} fuses PaLM-2~\citep{anil2023palm} and AudioLM with a joint vocabulary that can represent speech and text with discrete tokens.

\textbf{Discrete Indices from Other Techniques}: Apart from the aforementioned time-domain quantization, FreqTST~\citep{anonymous2023modeling} utilizes frequency spectrum as a common dictionary to discretize time series into frequency units with weights for downstream forecasting task. Chronos~\citep{ansari2024chronos} quantizes real-valued time
series into discrete bins, and optimizes existing language model architectures on these tokenized time series via the cross-entropy loss.

\textbf{Text Categories}: The second type of quantization converts numerical data into pre-defined text categories, which is primarily adopted in financial domain. For example, TDML~\citep{yu2023temporal} categorizes the weekly price fluctuations into 12 bins represented as ``D$i$'' or ``U$i$'', where ``D'' indicates a decrease in price and ``U'' means an increase, and $i=1, 2, 3, 4, 5, 5+$ represents the level of price change.

\subsection{Aligning}

\begin{figure}[t]
      \centering
      \begin{subfigure}[b]{0.45\textwidth}
          \centering
          \includegraphics[width=\textwidth]{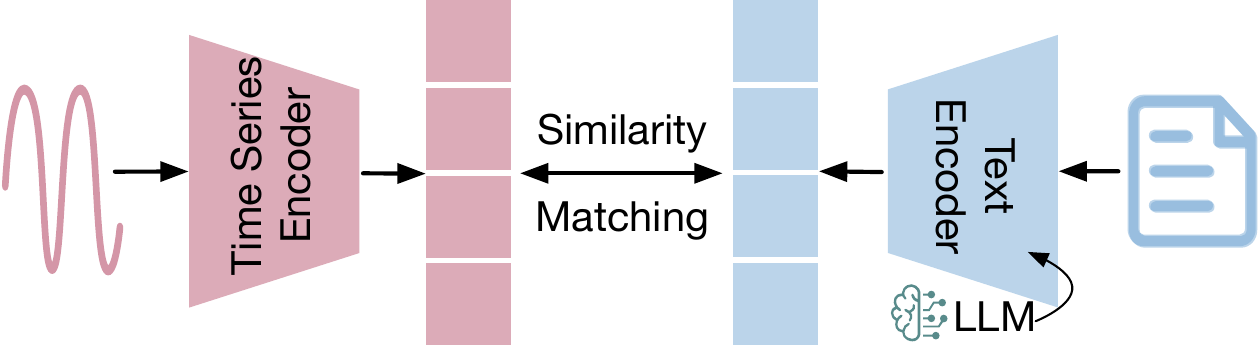}
          \caption{Aligning by similarity matching (Type one).}
      \end{subfigure}
      \begin{subfigure}[b]{0.45\textwidth}
          \centering
          \includegraphics[width=\textwidth]{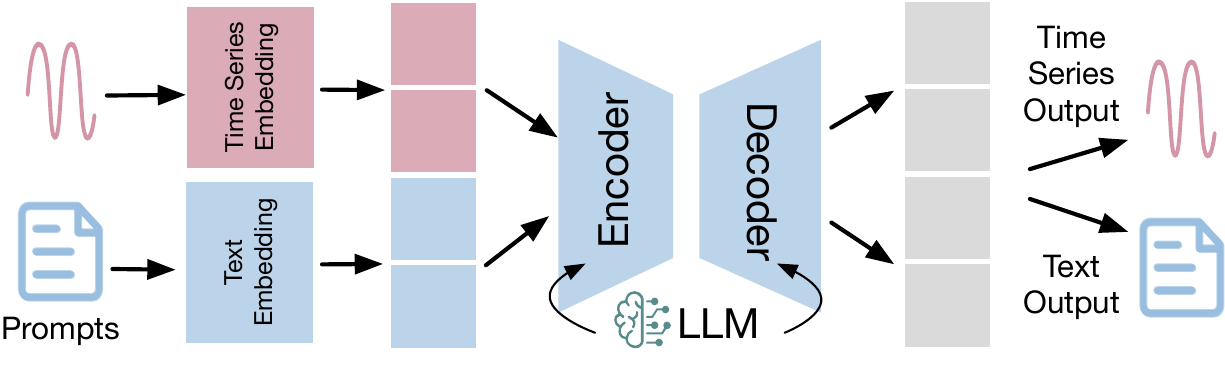}
          \caption{Aligning with large language models as backbones (Type two), where the output could be time series (e.g., forecasting) or text (e.g., EEG-to-text) depending on the downstream tasks.}
      \end{subfigure}
    \caption{Two types of aligning based methods.}
    \label{fig:alignment}
\end{figure}

The third type of works trains a separate encoder for time series, and aligns the encoded time series to the semantic space of language models. These works can be further categorized into two groups based on their specific aligning strategies, as illustrated in Figure~\ref{fig:alignment}.

\textbf{Similarity Matching through Contrastive Loss}: The first type of method aligns the time series embeddings with text embeddings through similarity matching, such as minimizing the contrastive loss: 
\begin{equation}
    \mathcal{L} = - \frac{1}{B} \sum_{i=1}^B \log \frac{\exp(\mathrm{sim}(g_{\phi}(\mathbf{x}_{si}), f_{\theta}(\mathbf{x}_{ti})))^{\frac{1}{\gamma}}}{\sum_{k=1}^B \exp(\mathrm{sim}(g_{\phi}(\mathbf{x}_{si}), f_{\theta}(\mathbf{x}_{tk})))^{\frac{1}{\gamma}}},
\end{equation}
where $B, \gamma$ represent batch size and temperature parameter that controls distribution concentrations, and $\mathrm{sim}$ represents similarity score, typically computed as inner product:
\begin{equation}
    \mathrm{sim}(g_{\phi}(\mathbf{x}_{si}), f_{\theta}(\mathbf{x}_{ti})) = \langle g_{\phi}(\mathbf{x}_{si}), f_{\theta}(\mathbf{x}_{ti}) \rangle.
\end{equation}

\renewcommand{\arraystretch}{1.5}
\begin{table*}[t]
\caption{Summary of five major categories of applying LLMs for time series analysis, including their respective subcategories, representative works, mathematical formulations, advantages and limitations. $q$ and $\mathbf{x}_v$ represent text-based quantization process and image data.}
\centering
{\small
\setlength{\tabcolsep}{1mm}{
\scalebox{0.96}{
\begin{tabular}{c|c|c|c|c|c}
\toprule
\rowcolor{gray!20} \textbf{Method} & \textbf{Subcategory} & \textbf{Representative Works} & \textbf{Equations} & \textbf{Advantages} & \textbf{Limitations} \\ \midrule
\multirow{2}{*}{Prompting} & Number-Agnostic & PromptCast~\citep{xue2022promptcast} & \multirow{2}{*}{$\mathbf{y}=f_{\theta}(\mathbf{x}_s, \mathbf{x}_t)$}& easy to implement; & lose semantics; \\ \cline{2-3}
& Number-Specific & LLMTime~\citep{gruver2023large} & & zero-shot capability & not efficient \\ \hline
\multirow{3}{*}{Quantization} & VQ-VAE & DeWave~\citep{duan2023dewave} & {\small $k_i = \arg\min_j \| g_{\phi}(\mathbf{x}_s)_i - \mathbf{c}_j \|_2$} & flexibility of & may require \\  \cline{2-3}
& K-Means & AudioLM~\citep{borsos2023audiolm} & $\mathbf{k} = [k_i]_{i=1}^{\frac{T}{S}}, \mathbf{y} = f_{\theta}(\mathbf{k}, \mathbf{x}_t)$ & index and time & two-stage \\ \cline{2-4}
& Text Categories & TDML~\citep{yu2023temporal} & $\mathbf{y}=f_{\theta}(q(\mathbf{x}_s), \mathbf{x}_t)$ & series conversion  & training \\ \hline
\multirow{3}{*}{Aligning} & \multirow{2}{*}{Similarity Match} & ETP~\citep{liu2023etp} & $\mathbf{y}=g_{\phi}(\mathbf{x}_s)$ & align semantics of & complicated \\ 
& & MATM~\citep{han2022empirical} & $\mathcal{L}=\mathrm{sim}(g_{\phi}(\mathbf{x}_s),f_{\theta}(\mathbf{x}_t))$ & different modalities; & design and \\ \cline{2-4}
& LLM Backbone & GPT4TS~\citep{zhou2023one} & $\mathbf{y}=f_{\theta}(g_{\phi}(\mathbf{x}_s), \mathbf{x}_t)$ & end-to-end training & fine-tuning  \\ \hline
Vision as & Paired Data & ImageBind~\citep{girdhar2023imagebind} & $\mathcal{L}=\mathrm{sim}(g_{\phi}(\mathbf{x}_s),h_{\psi}(\mathbf{x}_{v}))$  & additional visual & not hold \\ \cline{2-4}
Bridge & TS Plots as Images & \citet{wimmer2023leveraging} & $\mathbf{y}=h_{\psi}(\mathbf{x}_s)$ & knowledge &  for all data \\ \hline
\multirow{2}{*}{Tool} & Code & CTG++~\citep{zhong2023language} & $z=f_{\theta}(\mathbf{x}_t)$ & empower LLM & optimization \\ \cline{2-3}
& API & ToolLLM~\citep{qin2023toolllm} & $\mathbf{y}=z(\mathbf{x}_s)$ & with more abilities & not end-to-end \\ \bottomrule
\end{tabular}}}}
\label{tab:summary}
\end{table*}

For instance, ETP~\citep{liu2023etp,li2023frozen} integrates contrastive learning based pre-training to align electrocardiography (ECG) signals with textual reports. \citet{king2023multimodal} use similar contrastive framework to align 17 clinical measurements collected in Intensive Care Unit (ICU) to their corresponding clinical notes. TEST~\citep{sun2023test} uses contrastive learning to generate instance-wise,
feature-wise, and text-prototype-aligned time series embeddings to align with text embeddings. TENT~\citep{zhou2023tent} aligns text embeddings with IoT sensor signals through a unified semantic feature space using contrastive learning. JoLT~\citep{cai2023jolt} utilizes Querying Transformer (Q-Former)~\citep{li2023blip} optimized with contrastive loss to align the time series and text representations. 

\textbf{Similarity Matching through Other Losses}: Apart from contrastive loss, other loss functions are also employed to optimize similarity matching between time series embeddings and text embeddings. ECG-LLM~\citep{qiu2023transfer} aligns the distribution between
ECG and language embedding from ECG statements with an Optimal Transport based loss function to train an ECG report generation model. MTAM~\citep{han2022empirical} uses various aligning techniques, such as Canonical Correlation Analysis and Wasserstein Distance, as loss functions to align electroencephalography (EEG) features with their corresponding language descriptions.  

\textbf{LLMs as Backbones}: The second type of aligning method directly uses large language models as backbones following time series embedding layers. 
EEG-to-Text~\citep{wang2022open} feeds EEG embeddings to pre-trained BART for open vocabulary EEG-To-Text decoding and EEG-based sentiment classification.  
GPT4TS~\citep{zhou2023one} uses patching embeddings~\citep{nie2022time} as input to frozen pre-trained GPT-2 where the positional embedding layers and self-attention blocks are retained during time series fine-tuning. The method provides a unified framework for seven time series tasks, including few-shot or zero-shot learning.
Following GPT4TS, researchers further incorporated seasonal-trend decomposition (TEMPO~\citep{cao2023tempo}), two-stage fine-tuning (LLM4TS~\citep{chang2023llm4ts}), domain descriptions (UniTime~\citep{liu2023unitime}), graph attention mechanism (GATGPT~\citep{chen2023gatgpt}), and spatial-temporal embedding module (ST-LLM~\citep{liu2024spatial}).
Time-LLM~\citep{jin2023time} reprograms time series data into text prototypes as input to LLaMA-7B. It also provides natural language prompts such as domain expert knowledge and task instructions to augment input context. 
Lag-Llama~\citep{rasul2023lag} builds univariate probabilistic time series forecasting model based on LLaMA architecture. 
In the audio, speech and music domains, researchers have also designed dedicated encoders to embed speech (WavPrompt ~\citep{gao2022wavprompt}, Speech LLaMA~\citep{lakomkin2023end})), music (MU-LLaMA~\citep{liu2023music}), and general audio inputs (LTU~\citep{gong2023listen}, SALMONN~\citep{tang2023salmonn}), and feed the embeddings to large language models. 

\subsection{Vision as Bridge}

Time series data can be effectively interpreted or associated with visual representations, which align closer with textual data and have demonstrated successful integrations with large language models. Therefore, researchers have also leveraged vision modality as a bridge to connect time series with LLMs. 

\textbf{Paired Data}: ImageBind~\citep{girdhar2023imagebind} uses image-paired data to bind six modalities (images, text, audio, depth, thermal, and Inertial Measurement Unit (IMU) time series) and learn a joint embedding space, enabling new emergent capabilities. PandaGPT~\citep{su2023pandagpt} further combines the multimodal encoders from ImageBind and LLMs to enable visual and auditory instruction-following capabilities. IMU2CLIP~\citep{moon2022imu2clip} aligns IMU time series with video and text, by projecting them into the joint representation space of Contrastive Language-Image Pre-training (CLIP)~\citep{radford2021learning}. AnyMAL~\citep{moon2023anymal} builds upon IMU2CLIP by training a lightweight adapter to project the IMU embeddings into the text token embedding space of LLaMA-2-70B. It is also capable of transforming data from other
modalities, such as images, videos, audio, into the same text embedding space. 

\textbf{Physics Relationships}: IMUGPT~\citep{leng2023generating} generates IMU data from ChatGPT-augmented text descriptions. It first generates 3D human motion from text using pre-trained motion synthesis model T2M-GPT~\citep{zhang2023t2m}. Then it derives IMU data from 3D motion based on physics relationships of motion kinetics.

\textbf{Time Series Plots as Images}: CLIP-LSTM~\citep{wimmer2023leveraging} transforms stock market data into sequences of texts and images of price charts, and leverages pre-trained CLIP vision-language model to generate features for downstream forecasting. Insight Miner~\citep{zhang2023insight} converts time series windows into images using lineplot, and feeds images into vision language model LLaVA~\citep{liu2023visual} to generate time series trend descriptions. 

\subsection{Tool}
This type of method does not directly use large language models to process time series. Instead, it applies large language models to generate indirect tools $z(\cdot)$, such as code and API calls, to benefit time series related tasks. 

\textbf{Code}: CTG++~\citep{zhong2023language} applies GPT-4 to generate differentiable loss functions in a code format from text descriptions to guide the diffusion model to generate traffic trajectories. With this two-step translation, the LLM and diffusion model efficiently bridge the gap between user intent and traffic simulation.

\textbf{API Call}: ToolLLM~\citep{qin2023toolllm} introduces a general tool-use framework composed of data construction, model training, and evaluation. This framework includes API calls for time series tasks such as weather and stock forecasting.

\textbf{Text Domain Knowledge}: SHARE~\citep{zhang2023unleashing} exploits the shared structures in human activity label names and proposes a sequence-to-sequence structure to generate label names as token sequences to preserve the shared label structures. It applies GPT-4 to augment semantics of label names. GG-LLM~\citep{graule2023gg} leverages LLaMA-2 to encode world knowledge of common human behavioral patterns to predict human actions without further training. SCRL-LG~\citep{ding2023integrating} leverages LLaMA-7B as stock feature selectors to extract meaningful representations from news headlines, which are subsequently employed in reinforcement learning for precise feature alignments.

\renewcommand{\arraystretch}{1.5}
\begin{table*}[t]
\caption{Summary of representative time series and text multimodal datasets.}
\centering
{\small
\setlength{\tabcolsep}{1mm}{
\scalebox{0.85}{
\begin{tabular}{c|c|c|c|c}
\toprule
\rowcolor{gray!20} \textbf{Domain} & \textbf{Dataset} & \textbf{Size} & \textbf{Major Modalities} & \textbf{Task} \\ \midrule
\multirow{2}{*}{Internet of Things} & Ego4D\footnotemark~\citep{grauman2022ego4d} & $3,670$h data, $3.85$M narrations & text, IMU, video, audio, 3D & classification, forecasting \\ \cline{2-5}
& DeepSQA\footnotemark~\citep{xing2021deepsqa} & $25$h data, $91$K questions & text, imu & classification, question answering \\ \hline
\multirow{2}{*}{Finance} & PIXIU\footnotemark~\citep{xie2023pixiu} & $136$K instruction data & text, tables & 5 NLP tasks, forecasting \\ \cline{2-5}
& MoAT\footnotemark~\citep{anonymous2023moat} & $6$ datasets, $2$K timesteps in total  & text, time series & forecasting \\ \hline
\multirow{3}{*}{Healthcare} & Zuco 2.0\footnotemark~\citep{hollenstein2019zuco} & $739$ sentences & text, eye-tracking, EEG  & classification, text generation \\ \cline{2-5}
& PTB-XL\footnotemark~\citep{wagner2020ptb} & $60$h data, $71$ unique statements & text, ECG & classification \\ \cline{2-5}
& ECG-QA\footnotemark~\citep{oh2023ecg} & $70$ question templates & text, ECG & classification, question answering \\ \hline
Audio & OpenAQA-5M\footnotemark~\citep{gong2023listen} & $5.6$M (audio, question, answer) tuples & text, audio & tagging, classification \\ \cline{2-5}
Music & MusicCaps\footnotemark~\citep{agostinelli2023musiclm} & $5.5$K music clips & text, music & captioning, generation \\ \cline{2-5}
Speech & CommonVoice\footnotemark~\citep{ardila2019common} & $7,335$ speech hours in $60$ languages & text, speech & ASR, translation \\
\bottomrule
\end{tabular}}}}
\label{tab:dataset}
\end{table*}

\section{Comparison within the Taxonomy}

We compare the five categories of our taxonomy and provide general guidelines for which category to choose based on considerations of data, model, efficiency and optimization. 

\textbf{Data}: When no training data is available and the objective is to apply LLM for time series in an zero-shot fashion, it is preferable to use prompting-based methods. This is because direct prompting enables the utilization of pre-trained language models' inherent capabilities without fine-tuning. However, representing numbers as strings can diminish the semantic value intrinsically tied to numerical data. Therefore, with adequate training data, quantization or aligning-based methods become more advantageous.  As shown in Figure~\ref{fig:taxo}, these two categories are the most extensively studied ones in existing literature. Furthermore, if time series data can be interpreted or associated with visual representations, these representations can be incorporated to utilize the intrinsic knowledge embedded in the vision modality or pre-trained vision-language models. 

\textbf{Model}: Prompting and tool integration methods tend to apply billion-parameter models as they often apply off-the-self LLMs without architectural modifications. By contrast, aligning and quantization methods vary from million to billion-parameter models, depending on the specific application requirements and available computational resources.

\textbf{Efficiency}: Prompting-based methods are not efficient for numerical data with high precision, as well as multivariate time series as it requires transforming each dimension into separate univariate time series, resulting in extremely long input. They are also less efficient for long-term predictions due to the computational demands of generating long sequences. These methods are more effective when dealing with simple numerical data that is richly interwoven with textual information, such as opening and closing stock prices in financial news articles. By contrast, quantization and aligning methods are more efficient to handle long sequences, as time series are typically down-sampled or segmented into patches before feeding into large language models.

\textbf{Optimization}: Depending on the specific discretization technique, quantization based method may require a two-stage training process (such as first training the VQ-VAE model), which may result in sub-optimal performance compared with that achieved through end-to-end training in aligning methods. Using large language models as indirect tools empowers LLMs with more capabilities to manage numerical data, but also raises the level of complexity to optimize both LLMs and other components in an end-to-end fashion. Therefore, existing works of tool integration typically employ off-the-shelf LLMs without further fine-tuning. 

\addtocounter{footnote}{-9}
\footnotetext{\url{https://ego4d-data.org/}}
\addtocounter{footnote}{1}
\footnotetext{\url{https://github.com/nesl/DeepSQA}}
\addtocounter{footnote}{1}
\footnotetext{\url{https://github.com/chancefocus/PIXIU}}
\addtocounter{footnote}{1}
\footnotetext{\url{https://openreview.net/pdf?id=uRXxnoqDHH}}
\addtocounter{footnote}{1}
\footnotetext{\url{https://osf.io/2urht/}}
\addtocounter{footnote}{1}
\footnotetext{\url{https://physionet.org/content/ptb-xl/1.0.3/}}
\addtocounter{footnote}{1}
\footnotetext{\url{https://github.com/Jwoo5/ecg-qa}}
\addtocounter{footnote}{1}
\footnotetext{\url{https://github.com/YuanGongND/ltu}}
\addtocounter{footnote}{1}
\footnotetext{\url{https://www.kaggle.com/datasets/googleai/musiccaps}}
\addtocounter{footnote}{1}
\footnotetext{\url{https://commonvoice.mozilla.org/en/datasets}}
\section{Multimodal Datasets}

Applying LLMs for time series benefits from the availability of multimodal time series and text data. In this section, we introduce representative multimodal time series and text datasets organized by their respective domains (Table~\ref{tab:dataset}). We list additional multimodal datasets in our Github repository\footnote{\url{https://github.com/xiyuanzh/awesome-llm-time-series}}.

\noindent \textbf{Internet of Things (IoT)}: Human activity recognition is an important task in IoT domain, which identifies human activities given time series collected with IoT devices (such as IMU sensors). The corresponding text data are the labels or text descriptions of these activities. Ego4D~\citep{grauman2022ego4d} presents 3,670 hours of daily-life activity data across hundreds of scenarios, including household, outdoor, workplace, and leisure. The dataset is rich in modalities, including the IMU time series measurement, and dense temporally-aligned textual descriptions of the activities and object interactions, totaling 3.85 million sentences. Ego-Exo4D~\citep{grauman2023ego} further offers three kinds of paired natural language datasets including expert commentary, narrate-and-act descriptions provided by the participants, and atomic action descriptions similar as Ego4D. DeepSQA~\citep{xing2021deepsqa} presents a generalized Sensory Question Answering (SQA) framework to facilitate querying raw sensory data related to human activities using natural language.

\noindent \textbf{Finance}: PIXIU~\citep{xie2023pixiu} presents multi-task and multi-modal instruction tuning data in the financial domain with 136K data samples. It contains both financial natural language understanding and prediction tasks, and covers 9 datasets of multiple modalities such as text and time series. MoAT~\citep{anonymous2023moat} constructs multimodal datasets with textual information paired with time series for each timestep, such as news articles extracted with relevant keywords, mostly covering finance related domains such as fuel, metal, stock and bitcoin. 

\noindent \textbf{Healthcare}: Zuco 1.0~\citep{hollenstein2018zuco} and Zuco 2.0~\citep{hollenstein2019zuco} datasets contain simultaneous eye-tracking and EEG during natural reading and during annotation. PTB-XL~\citep{wagner2020ptb} offers comprehensive metadata regarding ECG annotated by expert cardiologists, covering information such as ECG reports, diagnostic statements, diagnosis likelihoods, and signal-specific properties. Based on PTB-XL, ECG-QA~\citep{oh2023ecg} introduces the first Question Answering (QA) dataset for ECG analysis, containg 70 question templates that cover a wide range of clinically relevant ECG topics.

\noindent \textbf{Audio/Music/Speech}: AudioSet~\citep{gemmeke2017audio} is a collection of 2 million 10-second audio clips excised from YouTube videos and labeled with the sounds that the clip contains from a set of 527 labels. OpenAQA-5M~\citep{gong2023listen} dataset consists of 1.9 million closed-ended and 3.7 million open-ended (audio, question, answer) tuples. MusicCaps~\citep{agostinelli2023musiclm} is a high-quality music caption dataset, including 5.5K music clips. MTG-Jamendo~\citep{bogdanov2019mtg} is a dataset with 55,000 audio songs in various languages. Libri-Light~\citep{kahn2020libri} is an English dataset encompassing 60,000 hours of speech data. CommonVoice~\citep{ardila2019common} is a multilingual speech dataset consisting of 7,335 validated hours in 60 languages.

These datasets offer valuable benchmarks for multimodal time series and text analysis. These contain both time series focused tasks, including classification, which is evaluated using accuracy and macro-F1 scores, and forecasting, which utilizes metrics such as MSE, MAE, RMSE, and MAPE, as well as NLP focused tasks such as captioning, question answering, and translation, assessed through BLEU, ROUGE, METEOR, and EM scores, among others.
\section{Challenges and Future Directions}
In this section, we introduce the challenges and promising future directions of applying LLMs for time series analysis.

\subsection{Theoretical Understanding}
Existing works empirically show the benefits of applying LLMs for time series analysis. For example, LIFT~\citep{dinh2022lift} empirically shows that language model fine-tuning can work for non-language tasks without changing the architecture or loss function; \citet{gurnee2023language} empirically show that LLMs learn linear representations of space and time across multiple scales that are robust to prompting variations. Despite these empirical findings, there remains a gap in theoretical understanding of how models, primarily trained on textual data, can effectively interpret numerical time series.
As a preliminary theoretical analysis, \citet{yun2019transformers} prove that Transformer models can universally approximate arbitrary continuous sequence-to-sequence functions on a compact domain.
Additionally, GPT4TS~\citep{zhou2023one} theoretically shows that such generic capability of large language models can be related to Principal Component Analysis (PCA), as minimizing the gradient with respect to the self-attention layer shares similarities with PCA.
Further investigations on the generalizability of LLMs on numerical data is essential to establish solid understanding of the synergy between LLMs and time series analysis. 

\subsection{Multimodal and Multitask Analysis}
Existing papers that apply LLMs for time series analysis mostly focus on single modality and single task at a time, such as forecasting, classification, text generation, and do not support simultaneous multimodal and multitask analysis. In computer vision and audio domains, models such as Unified-IO~\citep{lu2022unified} and UniAudio~\citep{yang2023uniaudio} have unified multiple input modalities into a sequence of discrete vocabulary tokens to support multiple tasks within a single transformer-based architecture.  
More research into leveraging LLMs for multimodal and multitask analysis would lead to more powerful time series foundation models.  

\subsection{Efficient Algorithms}
Time series, especially those that are multivariate or possess long history information may increase the computational complexity for existing large language models. Patching~\citep{nie2022time} has been a widely adopted strategy to improve performance as well as reduce complexity, but large patches may obscure the semantic information of time series and negatively impact the performance. Therefore, developing more efficient algorithms is crucial for facilitating large-scale time series analysis and enhancing interactions with end users.

\subsection{Combining Domain Knowledge}
Combining existing statistical domain knowledge with LLMs may further boost the model's capability for time series analysis. For example, TEMPO~\citep{cao2023tempo} applies time series seasonal-trend decomposition and treats decomposed components as different semantic inductive biases as input to the pre-trained transformer. FreqTST~\citep{anonymous2023modeling} leverages insights from the frequency domain by tokenizing single time series into frequency units with weights for downstream forecasting. Further incorporating domain knowledge, such as wavelet decomposition, auto-correlation analysis, and empirical mode decomposition may augment LLMs' capabilities in analyzing time series data. 

\subsection{Customization and Privacy}
Existing works on large language models and time series analysis typically train a global model for all end users. Training customized models for different users based on the global model may bring further benefits and flexibility. Another important consideration is privacy, especially as many time series data are collected in private settings for clinical purposes or smart home applications. As an initial attempt, FedAlign~\citep{zhang2023navigating} leverages federated learning frameworks and uses the expressive natural language class names as a common ground to align the latent spaces across different clients. Advancing research into model customization and user privacy preservation would broaden the scope and utility of LLM-empowered time series analysis.

\section{Conclusion}
We present the first survey that systematically analyzes the categorization of transferring knowledge from large language models for numerical time series analysis: direct prompting, time series quantization, aligning, the use of the vision modality to connect text and time series, and the integration of large language models with other analytical tools. For each category, we introduce their mathematical formulation, representative works, and compare their advantages and limitations. We also introduce representative multimodal text and time series datasets in various domains such as healthcare, IoT, finance, and audio. Concluding the paper, we outline the challenges and emerging directions for potential future research of LLM-empowered time series analysis.  

\section{Acknowledgements}
Our work is supported in part by ACE, one of the seven centers in JUMP 2.0, a Semiconductor Research Corporation (SRC) program sponsored by DARPA. Our work is also sponsored by NSF CAREER Award 2239440, NSF Proto-OKN Award 2333790, NIH Bridge2AI Center Program under award 1U54HG012510-01, Cisco-UCSD Sponsored Research Project, as well as generous gifts from Google, Adobe, and Teradata. Any opinions, findings, and conclusions or recommendations expressed herein are those of the authors and should not be interpreted as necessarily representing the views, either expressed or implied, of the U.S. Government. The U.S. Government is authorized to reproduce and distribute reprints for government purposes not withstanding any copyright annotation hereon.

\bibliographystyle{named}
\bibliography{ref}

\begin{thebibliography}{}

\bibitem[\protect\citeauthoryear{Agostinelli \bgroup \em et al.\egroup }{2023}]{agostinelli2023musiclm}
Andrea Agostinelli, Timo~I Denk, Zal{\'a}n Borsos, Jesse Engel, Mauro Verzetti, Antoine Caillon, Qingqing Huang, Aren Jansen, Adam Roberts, Marco Tagliasacchi, et~al.
\newblock Musiclm: Generating music from text.
\newblock {\em arXiv preprint arXiv:2301.11325}, 2023.

\bibitem[\protect\citeauthoryear{Anil \bgroup \em et al.\egroup }{2023}]{anil2023palm}
Rohan Anil, Andrew~M Dai, Orhan Firat, Melvin Johnson, Dmitry Lepikhin, Alexandre Passos, Siamak Shakeri, Emanuel Taropa, Paige Bailey, Zhifeng Chen, et~al.
\newblock Palm 2 technical report.
\newblock {\em arXiv preprint arXiv:2305.10403}, 2023.

\bibitem[\protect\citeauthoryear{Ansari \bgroup \em et al.\egroup }{2024}]{ansari2024chronos}
Abdul~Fatir Ansari, Lorenzo Stella, Caner Turkmen, Xiyuan Zhang, Pedro Mercado, Huibin Shen, Oleksandr Shchur, Syama~Sundar Rangapuram, Sebastian~Pineda Arango, Shubham Kapoor, et~al.
\newblock Chronos: Learning the language of time series.
\newblock {\em arXiv preprint arXiv:2403.07815}, 2024.

\bibitem[\protect\citeauthoryear{Ardila \bgroup \em et al.\egroup }{2019}]{ardila2019common}
Rosana Ardila, Megan Branson, Kelly Davis, Michael Henretty, Michael Kohler, Josh Meyer, Reuben Morais, Lindsay Saunders, Francis~M Tyers, and Gregor Weber.
\newblock Common voice: A massively-multilingual speech corpus.
\newblock {\em arXiv preprint arXiv:1912.06670}, 2019.

\bibitem[\protect\citeauthoryear{Bogdanov \bgroup \em et al.\egroup }{2019}]{bogdanov2019mtg}
Dmitry Bogdanov, Minz Won, Philip Tovstogan, Alastair Porter, and Xavier Serra.
\newblock The mtg-jamendo dataset for automatic music tagging.
\newblock ICML, 2019.

\bibitem[\protect\citeauthoryear{Borsos \bgroup \em et al.\egroup }{2023}]{borsos2023audiolm}
Zal{\'a}n Borsos, Rapha{\"e}l Marinier, Damien Vincent, Eugene Kharitonov, Olivier Pietquin, Matt Sharifi, Dominik Roblek, Olivier Teboul, David Grangier, Marco Tagliasacchi, et~al.
\newblock Audiolm: a language modeling approach to audio generation.
\newblock {\em IEEE/ACM Transactions on Audio, Speech, and Language Processing}, 2023.

\bibitem[\protect\citeauthoryear{Cai \bgroup \em et al.\egroup }{2023}]{cai2023jolt}
Yifu Cai, Mononito Goswami, Arjun Choudhry, Arvind Srinivasan, and Artur Dubrawski.
\newblock Jolt: Jointly learned representations of language and time-series.
\newblock In {\em Deep Generative Models for Health Workshop NeurIPS 2023}, 2023.

\bibitem[\protect\citeauthoryear{Cao \bgroup \em et al.\egroup }{2023}]{cao2023tempo}
Defu Cao, Furong Jia, Sercan~O Arik, Tomas Pfister, Yixiang Zheng, Wen Ye, and Yan Liu.
\newblock Tempo: Prompt-based generative pre-trained transformer for time series forecasting.
\newblock {\em arXiv preprint arXiv:2310.04948}, 2023.

\bibitem[\protect\citeauthoryear{Chang \bgroup \em et al.\egroup }{2023}]{chang2023llm4ts}
Ching Chang, Wen-Chih Peng, and Tien-Fu Chen.
\newblock Llm4ts: Two-stage fine-tuning for time-series forecasting with pre-trained llms.
\newblock {\em arXiv preprint arXiv:2308.08469}, 2023.

\bibitem[\protect\citeauthoryear{Chen \bgroup \em et al.\egroup }{2023}]{chen2023gatgpt}
Yakun Chen, Xianzhi Wang, and Guandong Xu.
\newblock Gatgpt: A pre-trained large language model with graph attention network for spatiotemporal imputation.
\newblock {\em arXiv preprint arXiv:2311.14332}, 2023.

\bibitem[\protect\citeauthoryear{Chowdhery \bgroup \em et al.\egroup }{2023}]{chowdhery2023palm}
Aakanksha Chowdhery, Sharan Narang, Jacob Devlin, Maarten Bosma, Gaurav Mishra, Adam Roberts, Paul Barham, Hyung~Won Chung, Charles Sutton, Sebastian Gehrmann, et~al.
\newblock Palm: Scaling language modeling with pathways.
\newblock {\em Journal of Machine Learning Research}, 24(240):1--113, 2023.

\bibitem[\protect\citeauthoryear{Chung \bgroup \em et al.\egroup }{2023}]{chung2023text}
Hyunseung Chung, Jiho Kim, Joon-myoung Kwon, Ki-Hyun Jeon, Min~Sung Lee, and Edward Choi.
\newblock Text-to-ecg: 12-lead electrocardiogram synthesis conditioned on clinical text reports.
\newblock In {\em ICASSP 2023-2023 IEEE International Conference on Acoustics, Speech and Signal Processing (ICASSP)}, pages 1--5. IEEE, 2023.

\bibitem[\protect\citeauthoryear{Devlin \bgroup \em et al.\egroup }{2018}]{devlin2018bert}
Jacob Devlin, Ming-Wei Chang, Kenton Lee, and Kristina Toutanova.
\newblock Bert: Pre-training of deep bidirectional transformers for language understanding.
\newblock {\em arXiv preprint arXiv:1810.04805}, 2018.

\bibitem[\protect\citeauthoryear{Ding \bgroup \em et al.\egroup }{2023}]{ding2023integrating}
Yujie Ding, Shuai Jia, Tianyi Ma, Bingcheng Mao, Xiuze Zhou, Liuliu Li, and Dongming Han.
\newblock Integrating stock features and global information via large language models for enhanced stock return prediction.
\newblock {\em arXiv preprint arXiv:2310.05627}, 2023.

\bibitem[\protect\citeauthoryear{Dinh \bgroup \em et al.\egroup }{2022}]{dinh2022lift}
Tuan Dinh, Yuchen Zeng, Ruisu Zhang, Ziqian Lin, Michael Gira, Shashank Rajput, Jy-yong Sohn, Dimitris Papailiopoulos, and Kangwook Lee.
\newblock Lift: Language-interfaced fine-tuning for non-language machine learning tasks.
\newblock {\em Advances in Neural Information Processing Systems}, 35:11763--11784, 2022.

\bibitem[\protect\citeauthoryear{Duan \bgroup \em et al.\egroup }{2023}]{duan2023dewave}
Yiqun Duan, Charles Zhou, Zhen Wang, Yu-Kai Wang, and Chin-teng Lin.
\newblock Dewave: Discrete encoding of eeg waves for eeg to text translation.
\newblock In {\em Thirty-seventh Conference on Neural Information Processing Systems}, 2023.

\bibitem[\protect\citeauthoryear{Gao \bgroup \em et al.\egroup }{2022}]{gao2022wavprompt}
Heting Gao, Junrui Ni, Kaizhi Qian, Yang Zhang, Shiyu Chang, and Mark Hasegawa-Johnson.
\newblock Wavprompt: Towards few-shot spoken language understanding with frozen language models.
\newblock {\em arXiv preprint arXiv:2203.15863}, 2022.

\bibitem[\protect\citeauthoryear{Gemmeke \bgroup \em et al.\egroup }{2017}]{gemmeke2017audio}
Jort~F Gemmeke, Daniel~PW Ellis, Dylan Freedman, Aren Jansen, Wade Lawrence, R~Channing Moore, Manoj Plakal, and Marvin Ritter.
\newblock Audio set: An ontology and human-labeled dataset for audio events.
\newblock In {\em 2017 IEEE international conference on acoustics, speech and signal processing (ICASSP)}, pages 776--780. IEEE, 2017.

\bibitem[\protect\citeauthoryear{Girdhar \bgroup \em et al.\egroup }{2023}]{girdhar2023imagebind}
Rohit Girdhar, Alaaeldin El-Nouby, Zhuang Liu, Mannat Singh, Kalyan~Vasudev Alwala, Armand Joulin, and Ishan Misra.
\newblock Imagebind: One embedding space to bind them all.
\newblock In {\em Proceedings of the IEEE/CVF Conference on Computer Vision and Pattern Recognition}, pages 15180--15190, 2023.

\bibitem[\protect\citeauthoryear{Gong \bgroup \em et al.\egroup }{2023}]{gong2023listen}
Yuan Gong, Hongyin Luo, Alexander~H Liu, Leonid Karlinsky, and James Glass.
\newblock Listen, think, and understand.
\newblock {\em arXiv preprint arXiv:2305.10790}, 2023.

\bibitem[\protect\citeauthoryear{Graule and Isler}{2023}]{graule2023gg}
Moritz~A Graule and Volkan Isler.
\newblock Gg-llm: Geometrically grounding large language models for zero-shot human activity forecasting in human-aware task planning.
\newblock {\em arXiv preprint arXiv:2310.20034}, 2023.

\bibitem[\protect\citeauthoryear{Grauman \bgroup \em et al.\egroup }{2022}]{grauman2022ego4d}
Kristen Grauman, Andrew Westbury, Eugene Byrne, Zachary Chavis, Antonino Furnari, Rohit Girdhar, Jackson Hamburger, Hao Jiang, Miao Liu, Xingyu Liu, et~al.
\newblock Ego4d: Around the world in 3,000 hours of egocentric video.
\newblock In {\em Proceedings of the IEEE/CVF Conference on Computer Vision and Pattern Recognition}, pages 18995--19012, 2022.

\bibitem[\protect\citeauthoryear{Grauman \bgroup \em et al.\egroup }{2023}]{grauman2023ego}
Kristen Grauman, Andrew Westbury, Lorenzo Torresani, Kris Kitani, Jitendra Malik, Triantafyllos Afouras, Kumar Ashutosh, Vijay Baiyya, Siddhant Bansal, Bikram Boote, et~al.
\newblock Ego-exo4d: Understanding skilled human activity from first-and third-person perspectives.
\newblock {\em arXiv preprint arXiv:2311.18259}, 2023.

\bibitem[\protect\citeauthoryear{Gruver \bgroup \em et al.\egroup }{2023}]{gruver2023large}
Nate Gruver, Marc Finzi, Shikai Qiu, and Andrew~Gordon Wilson.
\newblock Large language models are zero-shot time series forecasters.
\newblock {\em arXiv preprint arXiv:2310.07820}, 2023.

\bibitem[\protect\citeauthoryear{Gurnee and Tegmark}{2023}]{gurnee2023language}
Wes Gurnee and Max Tegmark.
\newblock Language models represent space and time.
\newblock {\em arXiv preprint arXiv:2310.02207}, 2023.

\bibitem[\protect\citeauthoryear{Han \bgroup \em et al.\egroup }{2022}]{han2022empirical}
William Han, Jielin Qiu, Jiacheng Zhu, Mengdi Xu, Douglas Weber, Bo~Li, and Ding Zhao.
\newblock An empirical exploration of cross-domain alignment between language and electroencephalogram.
\newblock {\em arXiv preprint arXiv:2208.06348}, 2022.

\bibitem[\protect\citeauthoryear{Hegselmann \bgroup \em et al.\egroup }{2023}]{hegselmann2023tabllm}
Stefan Hegselmann, Alejandro Buendia, Hunter Lang, Monica Agrawal, Xiaoyi Jiang, and David Sontag.
\newblock Tabllm: Few-shot classification of tabular data with large language models.
\newblock In {\em International Conference on Artificial Intelligence and Statistics}, pages 5549--5581. PMLR, 2023.

\bibitem[\protect\citeauthoryear{Hollenstein \bgroup \em et al.\egroup }{2018}]{hollenstein2018zuco}
Nora Hollenstein, Jonathan Rotsztejn, Marius Troendle, Andreas Pedroni, Ce~Zhang, and Nicolas Langer.
\newblock Zuco, a simultaneous eeg and eye-tracking resource for natural sentence reading.
\newblock {\em Scientific data}, 5(1):1--13, 2018.

\bibitem[\protect\citeauthoryear{Hollenstein \bgroup \em et al.\egroup }{2019}]{hollenstein2019zuco}
Nora Hollenstein, Marius Troendle, Ce~Zhang, and Nicolas Langer.
\newblock Zuco 2.0: A dataset of physiological recordings during natural reading and annotation.
\newblock {\em arXiv preprint arXiv:1912.00903}, 2019.

\bibitem[\protect\citeauthoryear{Hsu \bgroup \em et al.\egroup }{2021}]{hsu2021hubert}
Wei-Ning Hsu, Benjamin Bolte, Yao-Hung~Hubert Tsai, Kushal Lakhotia, Ruslan Salakhutdinov, and Abdelrahman Mohamed.
\newblock Hubert: Self-supervised speech representation learning by masked prediction of hidden units.
\newblock {\em IEEE/ACM Transactions on Audio, Speech, and Language Processing}, 29:3451--3460, 2021.

\bibitem[\protect\citeauthoryear{Jin \bgroup \em et al.\egroup }{2023a}]{jin2023time}
Ming Jin, Shiyu Wang, Lintao Ma, Zhixuan Chu, James~Y Zhang, Xiaoming Shi, Pin-Yu Chen, Yuxuan Liang, Yuan-Fang Li, Shirui Pan, et~al.
\newblock Time-llm: Time series forecasting by reprogramming large language models.
\newblock {\em arXiv preprint arXiv:2310.01728}, 2023.

\bibitem[\protect\citeauthoryear{Jin \bgroup \em et al.\egroup }{2023b}]{jin2023large}
Ming Jin, Qingsong Wen, Yuxuan Liang, Chaoli Zhang, Siqiao Xue, Xue Wang, James Zhang, Yi~Wang, Haifeng Chen, Xiaoli Li, et~al.
\newblock Large models for time series and spatio-temporal data: A survey and outlook.
\newblock {\em arXiv preprint arXiv:2310.10196}, 2023.

\bibitem[\protect\citeauthoryear{Kahn \bgroup \em et al.\egroup }{2020}]{kahn2020libri}
Jacob Kahn, Morgane Rivi{\`e}re, Weiyi Zheng, Evgeny Kharitonov, Qiantong Xu, Pierre-Emmanuel Mazar{\'e}, Julien Karadayi, Vitaliy Liptchinsky, Ronan Collobert, Christian Fuegen, et~al.
\newblock Libri-light: A benchmark for asr with limited or no supervision.
\newblock In {\em ICASSP 2020-2020 IEEE International Conference on Acoustics, Speech and Signal Processing (ICASSP)}, pages 7669--7673. IEEE, 2020.

\bibitem[\protect\citeauthoryear{King \bgroup \em et al.\egroup }{2023}]{king2023multimodal}
Ryan King, Tianbao Yang, and Bobak~J Mortazavi.
\newblock Multimodal pretraining of medical time series and notes.
\newblock In {\em Machine Learning for Health (ML4H)}, pages 244--255. PMLR, 2023.

\bibitem[\protect\citeauthoryear{Kreuk \bgroup \em et al.\egroup }{2022}]{kreuk2022audiogen}
Felix Kreuk, Gabriel Synnaeve, Adam Polyak, Uriel Singer, Alexandre D{\'e}fossez, Jade Copet, Devi Parikh, Yaniv Taigman, and Yossi Adi.
\newblock Audiogen: Textually guided audio generation.
\newblock {\em arXiv preprint arXiv:2209.15352}, 2022.

\bibitem[\protect\citeauthoryear{Lakomkin \bgroup \em et al.\egroup }{2023}]{lakomkin2023end}
Egor Lakomkin, Chunyang Wu, Yassir Fathullah, Ozlem Kalinli, Michael~L Seltzer, and Christian Fuegen.
\newblock End-to-end speech recognition contextualization with large language models.
\newblock {\em arXiv preprint arXiv:2309.10917}, 2023.

\bibitem[\protect\citeauthoryear{Lee \bgroup \em et al.\egroup }{2023}]{anonymous2023moat}
Geon Lee, Wenchao Yu, Wei Cheng, and Haifeng Chen.
\newblock Moat: Multi-modal augmented time series forecasting.
\newblock 2023.

\bibitem[\protect\citeauthoryear{Leng \bgroup \em et al.\egroup }{2023}]{leng2023generating}
Zikang Leng, Hyeokhyen Kwon, and Thomas Pl{\"o}tz.
\newblock Generating virtual on-body accelerometer data from virtual textual descriptions for human activity recognition.
\newblock {\em arXiv preprint arXiv:2305.03187}, 2023.

\bibitem[\protect\citeauthoryear{Lewis \bgroup \em et al.\egroup }{2019}]{lewis2019bart}
Mike Lewis, Yinhan Liu, Naman Goyal, Marjan Ghazvininejad, Abdelrahman Mohamed, Omer Levy, Ves Stoyanov, and Luke Zettlemoyer.
\newblock Bart: Denoising sequence-to-sequence pre-training for natural language generation, translation, and comprehension.
\newblock {\em arXiv preprint arXiv:1910.13461}, 2019.

\bibitem[\protect\citeauthoryear{Li \bgroup \em et al.\egroup }{2023a}]{li2023frozen}
Jun Li, Che Liu, Sibo Cheng, Rossella Arcucci, and Shenda Hong.
\newblock Frozen language model helps ecg zero-shot learning.
\newblock {\em arXiv preprint arXiv:2303.12311}, 2023.

\bibitem[\protect\citeauthoryear{Li \bgroup \em et al.\egroup }{2023b}]{anonymous2023modeling}
Junkai Li, Weizhi Ma, and Yang Liu.
\newblock Modeling time series as text sequence a frequency-vectorization transformer for time series forecasting.
\newblock 2023.

\bibitem[\protect\citeauthoryear{Li \bgroup \em et al.\egroup }{2023c}]{li2023blip}
Junnan Li, Dongxu Li, Silvio Savarese, and Steven Hoi.
\newblock Blip-2: Bootstrapping language-image pre-training with frozen image encoders and large language models.
\newblock {\em arXiv preprint arXiv:2301.12597}, 2023.

\bibitem[\protect\citeauthoryear{Liu \bgroup \em et al.\egroup }{2023a}]{liu2023etp}
Che Liu, Zhongwei Wan, Sibo Cheng, Mi~Zhang, and Rossella Arcucci.
\newblock Etp: Learning transferable ecg representations via ecg-text pre-training.
\newblock {\em arXiv preprint arXiv:2309.07145}, 2023.

\bibitem[\protect\citeauthoryear{Liu \bgroup \em et al.\egroup }{2023b}]{liu2023visual}
Haotian Liu, Chunyuan Li, Qingyang Wu, and Yong~Jae Lee.
\newblock Visual instruction tuning.
\newblock {\em arXiv preprint arXiv:2304.08485}, 2023.

\bibitem[\protect\citeauthoryear{Liu \bgroup \em et al.\egroup }{2023c}]{liu2023music}
Shansong Liu, Atin~Sakkeer Hussain, Chenshuo Sun, and Ying Shan.
\newblock Music understanding llama: Advancing text-to-music generation with question answering and captioning.
\newblock {\em arXiv preprint arXiv:2308.11276}, 2023.

\bibitem[\protect\citeauthoryear{Liu \bgroup \em et al.\egroup }{2023d}]{liu2023large}
Xin Liu, Daniel McDuff, Geza Kovacs, Isaac Galatzer-Levy, Jacob Sunshine, Jiening Zhan, Ming-Zher Poh, Shun Liao, Paolo Di~Achille, and Shwetak Patel.
\newblock Large language models are few-shot health learners.
\newblock {\em arXiv preprint arXiv:2305.15525}, 2023.

\bibitem[\protect\citeauthoryear{Liu \bgroup \em et al.\egroup }{2023e}]{liu2023unitime}
Xu~Liu, Junfeng Hu, Yuan Li, Shizhe Diao, Yuxuan Liang, Bryan Hooi, and Roger Zimmermann.
\newblock Unitime: A language-empowered unified model for cross-domain time series forecasting.
\newblock {\em arXiv preprint arXiv:2310.09751}, 2023.

\bibitem[\protect\citeauthoryear{Liu \bgroup \em et al.\egroup }{2024}]{liu2024spatial}
Chenxi Liu, Sun Yang, Qianxiong Xu, Zhishuai Li, Cheng Long, Ziyue Li, and Rui Zhao.
\newblock Spatial-temporal large language model for traffic prediction.
\newblock {\em arXiv preprint arXiv:2401.10134}, 2024.

\bibitem[\protect\citeauthoryear{Lu \bgroup \em et al.\egroup }{2022}]{lu2022unified}
Jiasen Lu, Christopher Clark, Rowan Zellers, Roozbeh Mottaghi, and Aniruddha Kembhavi.
\newblock Unified-io: A unified model for vision, language, and multi-modal tasks.
\newblock {\em arXiv preprint arXiv:2206.08916}, 2022.

\bibitem[\protect\citeauthoryear{Ma \bgroup \em et al.\egroup }{2023}]{ma2023survey}
Qianli Ma, Zhen Liu, Zhenjing Zheng, Ziyang Huang, Siying Zhu, Zhongzhong Yu, and James~T Kwok.
\newblock A survey on time-series pre-trained models.
\newblock {\em arXiv preprint arXiv:2305.10716}, 2023.

\bibitem[\protect\citeauthoryear{Mirchandani \bgroup \em et al.\egroup }{2023}]{mirchandani2023large}
Suvir Mirchandani, Fei Xia, Pete Florence, Brian Ichter, Danny Driess, Montserrat~Gonzalez Arenas, Kanishka Rao, Dorsa Sadigh, and Andy Zeng.
\newblock Large language models as general pattern machines.
\newblock {\em arXiv preprint arXiv:2307.04721}, 2023.

\bibitem[\protect\citeauthoryear{Moon \bgroup \em et al.\egroup }{2022}]{moon2022imu2clip}
Seungwhan Moon, Andrea Madotto, Zhaojiang Lin, Alireza Dirafzoon, Aparajita Saraf, Amy Bearman, and Babak Damavandi.
\newblock Imu2clip: Multimodal contrastive learning for imu motion sensors from egocentric videos and text.
\newblock {\em arXiv preprint arXiv:2210.14395}, 2022.

\bibitem[\protect\citeauthoryear{Moon \bgroup \em et al.\egroup }{2023}]{moon2023anymal}
Seungwhan Moon, Andrea Madotto, Zhaojiang Lin, Tushar Nagarajan, Matt Smith, Shashank Jain, Chun-Fu Yeh, Prakash Murugesan, Peyman Heidari, Yue Liu, et~al.
\newblock Anymal: An efficient and scalable any-modality augmented language model.
\newblock {\em arXiv preprint arXiv:2309.16058}, 2023.

\bibitem[\protect\citeauthoryear{Nie \bgroup \em et al.\egroup }{2022}]{nie2022time}
Yuqi Nie, Nam~H Nguyen, Phanwadee Sinthong, and Jayant Kalagnanam.
\newblock A time series is worth 64 words: Long-term forecasting with transformers.
\newblock {\em arXiv preprint arXiv:2211.14730}, 2022.

\bibitem[\protect\citeauthoryear{Oh \bgroup \em et al.\egroup }{2023}]{oh2023ecg}
Jungwoo Oh, Seongsu Bae, Gyubok Lee, Joon-myoung Kwon, and Edward Choi.
\newblock Ecg-qa: A comprehensive question answering dataset combined with electrocardiogram.
\newblock {\em arXiv preprint arXiv:2306.15681}, 2023.

\bibitem[\protect\citeauthoryear{OpenAI}{2023}]{OpenAI2023GPT4TR}
OpenAI.
\newblock Gpt-4 technical report.
\newblock {\em ArXiv}, abs/2303.08774, 2023.

\bibitem[\protect\citeauthoryear{Qin \bgroup \em et al.\egroup }{2023}]{qin2023toolllm}
Yujia Qin, Shihao Liang, Yining Ye, Kunlun Zhu, Lan Yan, Yaxi Lu, Yankai Lin, Xin Cong, Xiangru Tang, Bill Qian, et~al.
\newblock Toolllm: Facilitating large language models to master 16000+ real-world apis.
\newblock {\em arXiv preprint arXiv:2307.16789}, 2023.

\bibitem[\protect\citeauthoryear{Qiu \bgroup \em et al.\egroup }{2023}]{qiu2023transfer}
Jielin Qiu, William Han, Jiacheng Zhu, Mengdi Xu, Michael Rosenberg, Emerson Liu, Douglas Weber, and Ding Zhao.
\newblock Transfer knowledge from natural language to electrocardiography: Can we detect cardiovascular disease through language models?
\newblock {\em arXiv preprint arXiv:2301.09017}, 2023.

\bibitem[\protect\citeauthoryear{Rabanser \bgroup \em et al.\egroup }{2020}]{rabanser2020effectiveness}
Stephan Rabanser, Tim Januschowski, Valentin Flunkert, David Salinas, and Jan Gasthaus.
\newblock The effectiveness of discretization in forecasting: An empirical study on neural time series models.
\newblock {\em arXiv preprint arXiv:2005.10111}, 2020.

\bibitem[\protect\citeauthoryear{Radford \bgroup \em et al.\egroup }{2019}]{radford2019language}
Alec Radford, Jeffrey Wu, Rewon Child, David Luan, Dario Amodei, Ilya Sutskever, et~al.
\newblock Language models are unsupervised multitask learners.
\newblock {\em OpenAI blog}, 1(8):9, 2019.

\bibitem[\protect\citeauthoryear{Radford \bgroup \em et al.\egroup }{2021}]{radford2021learning}
Alec Radford, Jong~Wook Kim, Chris Hallacy, Aditya Ramesh, Gabriel Goh, Sandhini Agarwal, Girish Sastry, Amanda Askell, Pamela Mishkin, Jack Clark, et~al.
\newblock Learning transferable visual models from natural language supervision.
\newblock In {\em International conference on machine learning}, pages 8748--8763. PMLR, 2021.

\bibitem[\protect\citeauthoryear{Raffel \bgroup \em et al.\egroup }{2020}]{raffel2020exploring}
Colin Raffel, Noam Shazeer, Adam Roberts, Katherine Lee, Sharan Narang, Michael Matena, Yanqi Zhou, Wei Li, and Peter~J Liu.
\newblock Exploring the limits of transfer learning with a unified text-to-text transformer.
\newblock {\em The Journal of Machine Learning Research}, 21(1):5485--5551, 2020.

\bibitem[\protect\citeauthoryear{Rasul \bgroup \em et al.\egroup }{2023}]{rasul2023lag}
Kashif Rasul, Arjun Ashok, Andrew~Robert Williams, Arian Khorasani, George Adamopoulos, Rishika Bhagwatkar, Marin Bilo{\v{s}}, Hena Ghonia, Nadhir~Vincent Hassen, Anderson Schneider, et~al.
\newblock Lag-llama: Towards foundation models for time series forecasting.
\newblock {\em arXiv preprint arXiv:2310.08278}, 2023.

\bibitem[\protect\citeauthoryear{Rubenstein \bgroup \em et al.\egroup }{2023}]{rubenstein2023audiopalm}
Paul~K Rubenstein, Chulayuth Asawaroengchai, Duc~Dung Nguyen, Ankur Bapna, Zal{\'a}n Borsos, F{\'e}lix de~Chaumont Quitry, Peter Chen, Dalia~El Badawy, Wei Han, Eugene Kharitonov, et~al.
\newblock Audiopalm: A large language model that can speak and listen.
\newblock {\em arXiv preprint arXiv:2306.12925}, 2023.

\bibitem[\protect\citeauthoryear{Song \bgroup \em et al.\egroup }{2023}]{song2023bridge}
Shezheng Song, Xiaopeng Li, and Shasha Li.
\newblock How to bridge the gap between modalities: A comprehensive survey on multimodal large language model.
\newblock {\em arXiv preprint arXiv:2311.07594}, 2023.

\bibitem[\protect\citeauthoryear{Spathis and Kawsar}{2023}]{spathis2023first}
Dimitris Spathis and Fahim Kawsar.
\newblock The first step is the hardest: Pitfalls of representing and tokenizing temporal data for large language models.
\newblock {\em arXiv preprint arXiv:2309.06236}, 2023.

\bibitem[\protect\citeauthoryear{Su \bgroup \em et al.\egroup }{2023}]{su2023pandagpt}
Yixuan Su, Tian Lan, Huayang Li, Jialu Xu, Yan Wang, and Deng Cai.
\newblock Pandagpt: One model to instruction-follow them all.
\newblock {\em arXiv preprint arXiv:2305.16355}, 2023.

\bibitem[\protect\citeauthoryear{Sun \bgroup \em et al.\egroup }{2023}]{sun2023test}
Chenxi Sun, Yaliang Li, Hongyan Li, and Shenda Hong.
\newblock Test: Text prototype aligned embedding to activate llm's ability for time series.
\newblock {\em arXiv preprint arXiv:2308.08241}, 2023.

\bibitem[\protect\citeauthoryear{Talukder and Gkioxari}{2023}]{anonymous2023time}
Sabera~J Talukder and Georgia Gkioxari.
\newblock Time series modeling at scale: A universal representation across tasks and domains.
\newblock 2023.

\bibitem[\protect\citeauthoryear{Tang \bgroup \em et al.\egroup }{2023}]{tang2023salmonn}
Changli Tang, Wenyi Yu, Guangzhi Sun, Xianzhao Chen, Tian Tan, Wei Li, Lu~Lu, Zejun Ma, and Chao Zhang.
\newblock Salmonn: Towards generic hearing abilities for large language models.
\newblock {\em arXiv preprint arXiv:2310.13289}, 2023.

\bibitem[\protect\citeauthoryear{Touvron \bgroup \em et al.\egroup }{2023}]{touvron2023llama}
Hugo Touvron, Louis Martin, Kevin Stone, Peter Albert, Amjad Almahairi, Yasmine Babaei, Nikolay Bashlykov, Soumya Batra, Prajjwal Bhargava, Shruti Bhosale, et~al.
\newblock Llama 2: Open foundation and fine-tuned chat models.
\newblock {\em arXiv preprint arXiv:2307.09288}, 2023.

\bibitem[\protect\citeauthoryear{Van Den~Oord \bgroup \em et al.\egroup }{2017}]{van2017neural}
Aaron Van Den~Oord, Oriol Vinyals, et~al.
\newblock Neural discrete representation learning.
\newblock {\em Advances in neural information processing systems}, 30, 2017.

\bibitem[\protect\citeauthoryear{Wagner \bgroup \em et al.\egroup }{2020}]{wagner2020ptb}
Patrick Wagner, Nils Strodthoff, Ralf-Dieter Bousseljot, Dieter Kreiseler, Fatima~I Lunze, Wojciech Samek, and Tobias Schaeffter.
\newblock Ptb-xl, a large publicly available electrocardiography dataset.
\newblock {\em Scientific data}, 7(1):154, 2020.

\bibitem[\protect\citeauthoryear{Wang and Ji}{2022}]{wang2022open}
Zhenhailong Wang and Heng Ji.
\newblock Open vocabulary electroencephalography-to-text decoding and zero-shot sentiment classification.
\newblock In {\em Proceedings of the AAAI Conference on Artificial Intelligence}, volume~36, pages 5350--5358, 2022.

\bibitem[\protect\citeauthoryear{Wang \bgroup \em et al.\egroup }{2023a}]{wang2023viola}
Tianrui Wang, Long Zhou, Ziqiang Zhang, Yu~Wu, Shujie Liu, Yashesh Gaur, Zhuo Chen, Jinyu Li, and Furu Wei.
\newblock Viola: Unified codec language models for speech recognition, synthesis, and translation.
\newblock {\em arXiv preprint arXiv:2305.16107}, 2023.

\bibitem[\protect\citeauthoryear{Wang \bgroup \em et al.\egroup }{2023b}]{wang2023would}
Xinglei Wang, Meng Fang, Zichao Zeng, and Tao Cheng.
\newblock Where would i go next? large language models as human mobility predictors.
\newblock {\em arXiv preprint arXiv:2308.15197}, 2023.

\bibitem[\protect\citeauthoryear{Wen \bgroup \em et al.\egroup }{2022}]{wen2022transformers}
Qingsong Wen, Tian Zhou, Chaoli Zhang, Weiqi Chen, Ziqing Ma, Junchi Yan, and Liang Sun.
\newblock Transformers in time series: A survey.
\newblock {\em arXiv preprint arXiv:2202.07125}, 2022.

\bibitem[\protect\citeauthoryear{Wimmer and Rekabsaz}{2023}]{wimmer2023leveraging}
Christopher Wimmer and Navid Rekabsaz.
\newblock Leveraging vision-language models for granular market change prediction.
\newblock {\em arXiv preprint arXiv:2301.10166}, 2023.

\bibitem[\protect\citeauthoryear{Wu \bgroup \em et al.\egroup }{2023}]{wu2023bloomberggpt}
Shijie Wu, Ozan Irsoy, Steven Lu, Vadim Dabravolski, Mark Dredze, Sebastian Gehrmann, Prabhanjan Kambadur, David Rosenberg, and Gideon Mann.
\newblock Bloomberggpt: A large language model for finance.
\newblock {\em arXiv preprint arXiv:2303.17564}, 2023.

\bibitem[\protect\citeauthoryear{Xie \bgroup \em et al.\egroup }{2023a}]{xie2023wall}
Qianqian Xie, Weiguang Han, Yanzhao Lai, Min Peng, and Jimin Huang.
\newblock The wall street neophyte: A zero-shot analysis of chatgpt over multimodal stock movement prediction challenges.
\newblock {\em arXiv preprint arXiv:2304.05351}, 2023.

\bibitem[\protect\citeauthoryear{Xie \bgroup \em et al.\egroup }{2023b}]{xie2023pixiu}
Qianqian Xie, Weiguang Han, Xiao Zhang, Yanzhao Lai, Min Peng, Alejandro Lopez-Lira, and Jimin Huang.
\newblock Pixiu: A large language model, instruction data and evaluation benchmark for finance.
\newblock {\em arXiv preprint arXiv:2306.05443}, 2023.

\bibitem[\protect\citeauthoryear{Xing \bgroup \em et al.\egroup }{2021}]{xing2021deepsqa}
Tianwei Xing, Luis Garcia, Federico Cerutti, Lance Kaplan, Alun Preece, and Mani Srivastava.
\newblock Deepsqa: Understanding sensor data via question answering.
\newblock In {\em Proceedings of the International Conference on Internet-of-Things Design and Implementation}, pages 106--118, 2021.

\bibitem[\protect\citeauthoryear{Xue and Salim}{2022}]{xue2022promptcast}
Hao Xue and Flora~D Salim.
\newblock Promptcast: A new prompt-based learning paradigm for time series forecasting.
\newblock 2022.

\bibitem[\protect\citeauthoryear{Xue and Salim}{2023}]{xue2023utilizing}
Hao Xue and Flora~D Salim.
\newblock Utilizing language models for energy load forecasting.
\newblock In {\em Proceedings of the 10th ACM International Conference on Systems for Energy-Efficient Buildings, Cities, and Transportation}, pages 224--227, 2023.

\bibitem[\protect\citeauthoryear{Xue \bgroup \em et al.\egroup }{2022}]{xue2022leveraging}
Hao Xue, Bhanu~Prakash Voutharoja, and Flora~D Salim.
\newblock Leveraging language foundation models for human mobility forecasting.
\newblock In {\em Proceedings of the 30th International Conference on Advances in Geographic Information Systems}, pages 1--9, 2022.

\bibitem[\protect\citeauthoryear{Yang \bgroup \em et al.\egroup }{2023}]{yang2023uniaudio}
Dongchao Yang, Jinchuan Tian, Xu~Tan, Rongjie Huang, Songxiang Liu, Xuankai Chang, Jiatong Shi, Sheng Zhao, Jiang Bian, Xixin Wu, et~al.
\newblock Uniaudio: An audio foundation model toward universal audio generation.
\newblock {\em arXiv preprint arXiv:2310.00704}, 2023.

\bibitem[\protect\citeauthoryear{Yu \bgroup \em et al.\egroup }{2023}]{yu2023temporal}
Xinli Yu, Zheng Chen, Yuan Ling, Shujing Dong, Zongyi Liu, and Yanbin Lu.
\newblock Temporal data meets llm--explainable financial time series forecasting.
\newblock {\em arXiv preprint arXiv:2306.11025}, 2023.

\bibitem[\protect\citeauthoryear{Yun \bgroup \em et al.\egroup }{2019}]{yun2019transformers}
Chulhee Yun, Srinadh Bhojanapalli, Ankit~Singh Rawat, Sashank~J Reddi, and Sanjiv Kumar.
\newblock Are transformers universal approximators of sequence-to-sequence functions?
\newblock {\em arXiv preprint arXiv:1912.10077}, 2019.

\bibitem[\protect\citeauthoryear{Zeghidour \bgroup \em et al.\egroup }{2021}]{zeghidour2021soundstream}
Neil Zeghidour, Alejandro Luebs, Ahmed Omran, Jan Skoglund, and Marco Tagliasacchi.
\newblock Soundstream: An end-to-end neural audio codec.
\newblock {\em IEEE/ACM Transactions on Audio, Speech, and Language Processing}, 30:495--507, 2021.

\bibitem[\protect\citeauthoryear{Zhang \bgroup \em et al.\egroup }{2023a}]{zhang2023speechgpt}
Dong Zhang, Shimin Li, Xin Zhang, Jun Zhan, Pengyu Wang, Yaqian Zhou, and Xipeng Qiu.
\newblock Speechgpt: Empowering large language models with intrinsic cross-modal conversational abilities.
\newblock {\em arXiv preprint arXiv:2305.11000}, 2023.

\bibitem[\protect\citeauthoryear{Zhang \bgroup \em et al.\egroup }{2023b}]{zhang2023t2m}
Jianrong Zhang, Yangsong Zhang, Xiaodong Cun, Shaoli Huang, Yong Zhang, Hongwei Zhao, Hongtao Lu, and Xi~Shen.
\newblock T2m-gpt: Generating human motion from textual descriptions with discrete representations.
\newblock {\em arXiv preprint arXiv:2301.06052}, 2023.

\bibitem[\protect\citeauthoryear{Zhang \bgroup \em et al.\egroup }{2023c}]{zhang2023navigating}
Jiayun Zhang, Xiyuan Zhang, Xinyang Zhang, Dezhi Hong, Rajesh~K. Gupta, and Jingbo Shang.
\newblock {Navigating Alignment for Non-identical Client Class Sets: A Label Name-Anchored Federated Learning Framework}.
\newblock In {\em Proceedings of the 29th {ACM} {SIGKDD} Conference on Knowledge Discovery and Data Mining}. {ACM}, aug 2023.

\bibitem[\protect\citeauthoryear{Zhang \bgroup \em et al.\egroup }{2023d}]{zhang2023unleashing}
Xiyuan Zhang, Ranak~Roy Chowdhury, Jiayun Zhang, Dezhi Hong, Rajesh~K. Gupta, and Jingbo Shang.
\newblock Unleashing the power of shared label structures for human activity recognition.
\newblock In {\em Proceedings of the 32nd ACM International Conference on Information and Knowledge Management}, CIKM '23, page 3340–3350. Association for Computing Machinery, 2023.

\bibitem[\protect\citeauthoryear{Zhang \bgroup \em et al.\egroup }{2023e}]{zhang2023insight}
Yunkai Zhang, Yawen Zhang, Ming Zheng, Kezhen Chen, Chongyang Gao, Ruian Ge, Siyuan Teng, Amine Jelloul, Jinmeng Rao, Xiaoyuan Guo, et~al.
\newblock Insight miner: A time series analysis dataset for cross-domain alignment with natural language.
\newblock In {\em NeurIPS 2023 AI for Science Workshop}, 2023.

\bibitem[\protect\citeauthoryear{Zhang \bgroup \em et al.\egroup }{2023f}]{zhang2023large}
Zheng Zhang, Hossein Amiri, Zhenke Liu, Andreas Z{\"u}fle, and Liang Zhao.
\newblock Large language models for spatial trajectory patterns mining.
\newblock {\em arXiv preprint arXiv:2310.04942}, 2023.

\bibitem[\protect\citeauthoryear{Zhong \bgroup \em et al.\egroup }{2023}]{zhong2023language}
Ziyuan Zhong, Davis Rempe, Yuxiao Chen, Boris Ivanovic, Yulong Cao, Danfei Xu, Marco Pavone, and Baishakhi Ray.
\newblock Language-guided traffic simulation via scene-level diffusion.
\newblock {\em arXiv preprint arXiv:2306.06344}, 2023.

\bibitem[\protect\citeauthoryear{Zhou \bgroup \em et al.\egroup }{2023a}]{zhou2023one}
Tian Zhou, Peisong Niu, Xue Wang, Liang Sun, and Rong Jin.
\newblock One fits all: Power general time series analysis by pretrained lm.
\newblock {\em arXiv preprint arXiv:2302.11939}, 2023.

\bibitem[\protect\citeauthoryear{Zhou \bgroup \em et al.\egroup }{2023b}]{zhou2023tent}
Yunjiao Zhou, Jianfei Yang, Han Zou, and Lihua Xie.
\newblock Tent: Connect language models with iot sensors for zero-shot activity recognition.
\newblock {\em arXiv preprint arXiv:2311.08245}, 2023.

\end{thebibliography}

\end{document}